\documentclass{article}

\usepackage{arxiv}

\usepackage[utf8]{inputenc} % allow utf-8 input
\usepackage[T1]{fontenc}    % use 8-bit T1 fonts
\usepackage{hyperref}       % hyperlinks
\usepackage{url}            % simple URL typesetting
\usepackage{booktabs}       % professional-quality tables
\usepackage{amsfonts}       % blackboard math symbols
\usepackage{nicefrac}       % compact symbols for 1/2, etc.
\usepackage{microtype}      % microtypography
\usepackage{lipsum}

\usepackage{verbatim}
\usepackage{graphicx}

\title{Evolving Self-supervised Neural Networks: Autonomous Intelligence from Evolved Self-teaching}

\author{
  Nam Le \\
  Natural Computing Research \& Applications Group\\
  University College Dublin\\
  Dublin, Ireland \\
  \texttt{nam.lehai@ucdconnect.ie} \\
}

\begin{document}
\maketitle

\begin{abstract}
This paper presents a technique called evolving self-supervised neural networks -- neural networks that can teach themselves, intrinsically motivated, without external supervision or reward. The proposed method presents some sort-of \textit{paradigm shift}, and differs greatly from both traditional gradient-based learning and evolutionary algorithms in that it combines the metaphor of evolution and learning, more specifically self-learning, together, rather than treating these phenomena alternatively. I simulate a multi-agent system in which neural networks are used to control autonomous foraging agents with little domain knowledge. Experimental results show that only evolved self-supervised agents can demonstrate some sort of intelligent behaviour, but not evolution or self-learning alone. Indications for future work on evolving intelligence are also presented.
\end{abstract}

% keywords can be removed

\section{Introduction}

Much of the recent neural network (NN) research has focused on the fields of `deep learning' (DL) \cite{Lecun2015}, \cite{Schmidhuber14} or `deep reinforcement learning' (DRL) \cite{Mnih2015}. Neuroevolution (NE) and gradient-based learning methods have been used alternatively to train the weights of a NN. Metaphorically, the latter approach is inspired by learning-from-experience in which backpropagation \cite{Rumelhart1986} is the most common method to train the weights of a NN. This approach has proven to be impressively effective in supervised learning, especially in some reinforcement learning tasks so that there has been some strong claim that we can build human-intelligence by learning from scratch as \textit{blank-slates}, without any \textit{innate} knowledge \cite{Mnih2015}, \cite{Silver2017}. Despite these impressive successes, there has been some objection to this line of thought that we are still very far from the so-called \textit{Artificial General Intelligence} (AGI) and innate knowledge is necessary to reach AGI \cite{Marcus-Innate}.

On another side, the NE approach uses a family of Evolutionary Algorithms which take inspiration from the fact that the natural brain is also the product of an evolutionary process. This approach was mostly developed in isolation from the gradient-based machine learning (ML) community, and has recently attracted interest from the ML community with several notable successes by OpenAI \cite{Evo-OpenAI-Salimans} and Uber AI Labs \cite{UberAI-Stanley-DeepNE}.

Way back to the metaphorical level, in nature evolution and learning are not alternative but complementary. It has been said that intelligent behaviour produced by animal agents (including humans) is the product of the interplay between both nature and nurture, or innate and learned \cite{Pinker-Blankslate}, \cite{Sasaki2016}, \cite{SAMUELS2004-Innate}. Learning as a lifetime adaptation can change the evolutionary pathway and evolution can provide a base for learning to take place. Evolution can be thought of as not producing an absolute solution, but creating a learner (with or without brain) which then learn to find the solution more effectively than learning as blank-slates. This is greatly different from the way a number of EA researcher have adopted the idea of evolution to seek for a solution to optimization problems (including the one of optimizing neural networks).

This paper aims at two main things. First, I present a technique called evolving self-supervised neural networks, or neural networks that can teach themselves \textit{intrinsically} without external supervisory signals. This network takes advantage of both evolutionary and gradient-descent learning to seek for its optimal weights. This is an important aspect of this contribution. Second, I simulate a multi-agent foraging world to test the performance of my proposed method.

Although there should possibly be a mixture of flavour, due to scope limit the reader is invited to have a look at \cite{Schmidhuber14} for an overview of neural network and deep learning, and \cite{Stanley2019} for a survey of neuroevolution approach to deep learning. In the remainder of this contribution, I initially present some prior research relating to learning and evolution in nature and in evolving neural networks. I then described the proposed technique and simulation undertaken. Results are then discussed, and concluded.

\section{Related Work}
The classic debate between \textit{nature vs nurture}, or \textit{innate vs learned} takes the separation between \textit{evolution vs learning} in explaining behaviour \cite{Sasaki2016}, \cite{Pinker-Blankslate}. Recent studies from Evolutionary Psychology \cite{barkow1995adapted}, \cite{Pinker-how-mind-works} have shown that neither of these extremes can give the full account for the behaviour of living organisms, especially when it comes to explain the behaviour of humans. This is to say, our species are not totally \textit{blank-slates}, and not totally \textit{instinctual} too. Most of our adaptive behaviours are the product of the interaction between nature and nurture, or innate and learned, or evolution and learning.

There existed an intriguing idea called \textbf{the Baldwin Effect} presenting how a behaviour first learned can then become innate or partially innate in a Darwinian framework, not Lamarckian\footnote{Lamarckism says acquired behaviour is directly transmitted into the genome of the offspring, which has not been accepted in the mainstream of evolutionary biology.} \cite{Baldwin:1896}, \cite{Simpson:1953}. Interestingly, this idea was largely neglected in the mainstream of biology and psychology for over 100 years, and just gained more attention since the classic paper by the British Cognitive Scientist Geoffrey Hinton and his colleague entitled \textit{How Learning Can Guide Evolution} in which an instance of the Baldwin Effect is demonstrated in a computer simulation \cite{Hinton:1987}. This paper stimulated a number of important follow-on studies including \cite{Nolfi:1994}, \cite{Mayley:1997}, \cite{Harvey:1996:NFE:1326720.1326727}. All of these studies attempted to show that individual learning can promote the population-based search process of evolution.

Learning and evolution in neural network learning has been studied in several papers following the original work of Hinton and Nowlan \cite{Hinton:1987}. Notable studies include  \cite{Keesing1990EvolutionAL} in which the authors used a genetic algorithm to evolve the initial weights of a digit classifier neural network which then can be learned by backpropagation. They found that if the amount of learning is used properly, learning can take advantage of starting weights produced by evolution to further the classification performance.

Nolfi and his colleagues made a simulation of \textit{animats}, or robots, controlled by neural networks situated in a grid-world, with discrete state and action spaces \cite{Nolfi:1994}. Each agent lives in its own copy of the world, hence no mutual interaction. The evolutionary task is to evolve action strategies to collect food effectively, while each agent learns to predict the sensory inputs to neural networks for each time step. Learning was implemented using backpropagation based on the error between the actual and the predicted sensory inputs to update the weights of a neural network. It was shown that learning to predict can enhance the evolutionary search, hence increasing the performance of the robot.

Scientists at DeepMind recently also used the same idea of the Baldwin Effect, using a genetic algorithm to evolve the initial weights for deep neural networks \cite{Deepmind-Baldwin}. By combining the advantage of searching over a vast distribution of weights of evolutionary search and the exploitative power of a gradient descent learning, they reported a \textit{meta-learner} that can solve a multiple tasks including regression and physical robot environments. This result is an indication to create meta-learning, another step towards AGI.

Generally learning in neural networks can be thought of as part of neural plasticity. There have been some other ideas, like evolving local learning rules to update the weights \cite{Bengio-rule-learning}, evolution of neuromodulation which facilitates the information transfer between neurons in hopes of creating meta-learning \cite{DOYA-meta-neuromodulation}. Please refer to \cite{Soltoggio2018} for more recent studies on evolving plastic neural networks. In short, most of the work use \textit{disembodied} and \textit{unsituated} neural networks in single-agent environment, having no mutual interaction as they solve their own problems, having no effect on other's performance.

In this paper, I adopt a different method called evolving self-supervised neural networks (SSNNs) (which was presented in \cite{LeNam:AISB:2019}). This technique differs from what has been presented so far in this section in that a SSNN consists of two separate network modules, namely action module and reinforcement module. The agent learns as the reinforcement module produces a signal to update the weights of the action module. This happens inside the agent whenever the agent moves and senses the environment, in need of no external \textit{supervision or reward}, thus intrinsically motivated. This technique also differs greatly from traditional supervised learning in which a learning machine is provided with labels, and from reinforcement learning in which a reward function is set by the system engineer.

Unlike previous work, I simulate a situated multi-agent system. Autonomous multi-agent learning has been stated as a step toward AGI \cite{Stone:2007}, \cite{OpenAI-Multiagent}. \textit{Situatedness} here means the intelligent behaviour of an agent is coupled with its environment: the agent controls its own sensory inputs and actions in the environment. Please refer to \cite{ANDERSON2003-embodied} for a review of situated and embodied approach to cognitive science and AI -- an inevitable part of general intelligence, including that of humans.

Our simulations are described in the following section. We shall be seeing how evolution and self-supervised learning interact to create more adaptive autonomous agents.

\section{Method}

\subsection{MiniWorld: Agents and Substances}
\begin{figure*}[t]
\begin{center}
\includegraphics[width=4in]{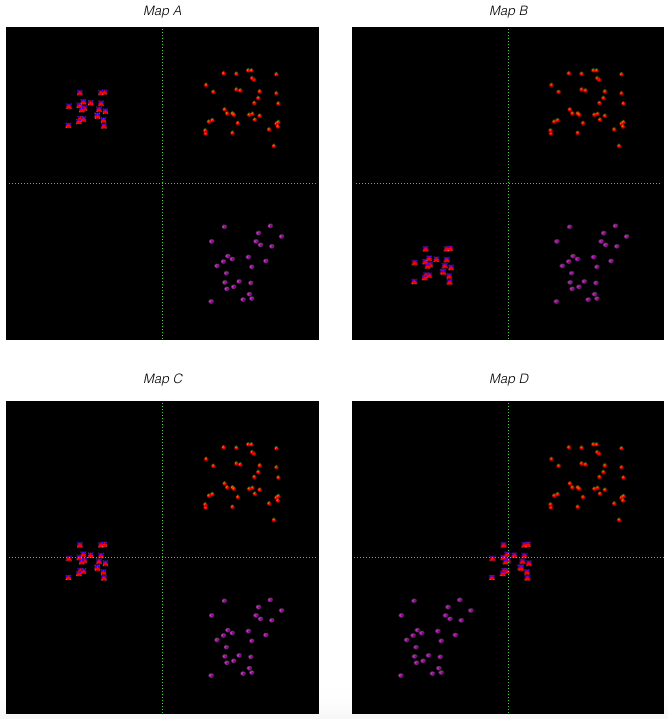}
\caption{MiniWorld -- The environment of agents and substances, 640x640. Denote $w$ and $h$ the width and the height of MiniWorld.  In each map, initially all agents are located around the vicinity of radius 40 (4 times the size of an agent) around one central point: ($w/4$, $h/4$), ($w/4$, $3h/4$), ($w/4$, $h/2$), and ($w/2$, $h/2$) for map A, B, C, and D, respectively. The food in all maps has horizontal and vertical dimensions randomly chosen in ranges ($5w/8$, $7w/8$) and ($h/8$, $3h/8$), respectively. Poisonous substances in 3 maps A, B, C have their horizontal and vertical dimensions randomly chosen in ranges ($5w/8$, $7w/8$) and ($5h/8$, $7h/8$), accordingly. The poison in map D has its horizontal and vertical dimensions randomly sampled from ranges ($w/8$, $3w/8$) and ($5h/8$, $7h/8$), respectively. Food has its colour representation as either [1,1,1,0] or [1,1,0,1]. Poison has its colour representation as either [0,1,1,1] or [1,0,1,1]. This creates a sort-of simple noisy sensory information, and may make it harder in discriminating between food and poison since two substances of different types may share a portion of colour representation.}
\label{fig:env}
\end{center}
\end{figure*}

Suppose that 20 agents situate in a continuous square 640x640 2D-world, called \textbf{MiniWorld}, which contains 60 substances of two types: food and poison. The multi-agent has to maximise its energy. Agents and substances are initially located at separate regions in MiniWorld depending on the world map. I propose 4 world maps (A, B, C, and D) in my simulations with details described in Figure \ref{fig:env}\footnote{Please note that dim green lines are sketched only for the purpose of visualising the world map of the food and the agent, MiniWorld is a continuous world, not grid-like. One property of MiniWorld is it has no strict boundary, and I implement the so-called \textit{toroidal} -- this means when an agent moves beyond an edge, it appears in the opposite edge.}. Each substance is represented by a square image of size 10x10. For simplicity I use a small binary bit sequence to represent the colour of the substance as described in Figure \ref{fig:env}. 

Each agent in MiniWorld has a squared body of size 10x10. When an agent's body happens to collide with a substance, the substance is eaten and another piece of the same type randomly spawn in the same region but at a different location. The energy level of the agent increases or decreases by 1 if the agent eats food or toxin, respectively. The agent body somehow affects how the agent senses and acts in MiniWorld.

\begin{figure*}[t]
\begin{center}
\includegraphics[width=2.5in]{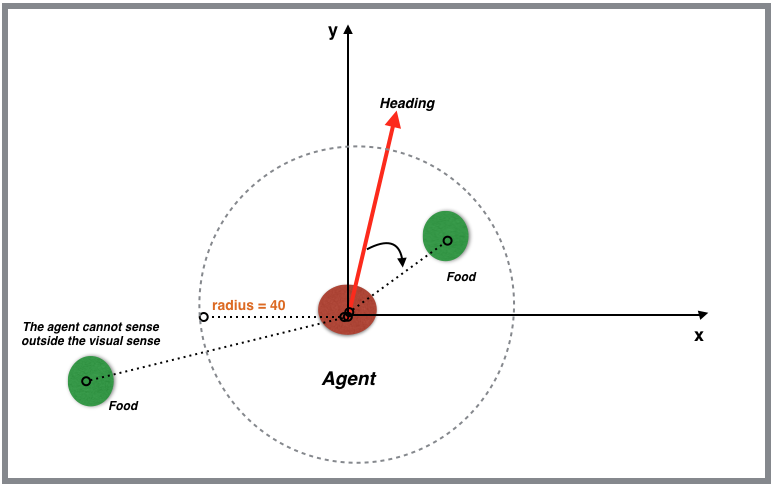}
\caption{Agent and substances in MiniWorld. Each agent has a visual range of radius 40. An agent takes as inputs \textbf{7 sensory information}. The first three bits (left, front, right) are set to 0 or 1 depending on whether the substance appears (in the left, front, and right) or not. Let $\theta$ (in degree) be the angle between the agent and the substance in its visual sense. An agent determines whether a food appears in its left, front, or right location in its visual range be the following rule: Right if 15 $<$ $\theta$ $<$ 45; Front if $\theta < 15$  or $\theta > 345$; and Left if $315 < \theta <345$. The last 4 bits contain information about the colour representation of the substance that \textbf{the agent senses in its visual range}. The dynamic and overlapping of substance colour representation makes the task of survival even harder for the agent since it has to deal with some sort of \textbf{noisy} sensory inputs. Indeed, the agent has to discriminate between food and poison whenever it encounters a substance to maximise its in-take.
}
\label{fig:agent-food}
\end{center}
\end{figure*}

Every agent has a heading (in principle) of movement in the environment. To be more controllable, all agents are initialised with a horizontal heading (i.e. with 0 degree), not a random orientation. This deliberate setting is for purpose of our world maps. A visualisation of an agent and its relationship with substance in the environment is shown in Figure \ref{fig:agent-food}.

%In our simulation, we assume that every agent has a \textit{priori} ability to sense the angle between its current heading and the substance if appearing in its visual range. 
  %If there is no substance appearing in its visual range, these three sensory inputs are all set to 0. If there is a substance appearing on the left (front, or right), the left (front, or right) sensor is set to 1; otherwise, the sensor is 0. 

The default velocity (or speed) of each agent is 1. Every agent has 5 basic movements: Turn left by 20 degrees and move, move forward by double speed, turn right by 20 degrees and move, move in the reverse direction, and stop at the current location. For now, these rules are pre-defined by the system designer of MiniWorld \footnote{We can imagine the perfect scenario like if an agent sees a food in front, it doubles the speed and move forward to catch the food. If the agent sees the food on the left (right), it should turn to the left (right) and move forward to the food particle. In cases in which the agent \textit{interprets} that the substance is poisonous, it would be likely to turn and move in the reverse direction, or \textbf{avoidance} action. When an agent is not sure about its current circumstance, it may pause to think more}.

\textbf{Note} that in our game, the agent has been equipped with very \textbf{little domain knowledge} about the world (e.g. state of the world, its current location or the relative distance between its current location to the source of substance). The agent is clearly \textbf{autonomous}. It has to experience the world on its own and acquire correct behaviour to find where the substance is. Moreover, the agent has \textbf{no priori-knowledge} about whether a substance it sees is food or poison. It has to \textbf{implicitly} develop an ability to discriminate between food and poison, through which it produces correct action (to approach or to avoid). Unlike reinforcement learning, no immediate reward is engineered in MiniWorld. The neural network method being proposed equips the agent with an intrinsic learning ability, which has been said a key to autonomous intelligence \cite{NIPS2004-Barto-Intrinsic}.

\subsection{The Neural Network Controller}
\label{sec:ann-controller}
Each agent is controlled by a fully-connected neural network to determine its movements in the environment as shown in Figure \ref{fig:neural-controller} a). All neurons except the inputs use a sigmoidal activation function. All connections (or synaptic strengths) are initialised as Gaussian(0, 1). These weights are first initialised as \textit{innate}, but also have the potential to change during the lifetime of that agent. Please note that what an agent decides to do changes the world the agent lives in, changing the next sensory information it receives, hence the next behaviour. This forms a sensory-motor dynamics and a neural network acts as a \textbf{situated cognitive module} having the role to guide an agent to behave effectively and adaptively. This is \textit{situated cognition} or \textit{situated intelligence} \cite{ANDERSON2003-embodied}.

In fact, the neural architecture as shown in Figure \ref{fig:neural-controller}a) has no ability to supervise, or to teach itself. I extend this architecture to allow for \textbf{self-supervised} learning agents.

\subsection{The Self-supervised Neural Architecture}
\label{sec:self-taught-ann}
\begin{figure}[t]
\begin{center}
\includegraphics[width=5in]{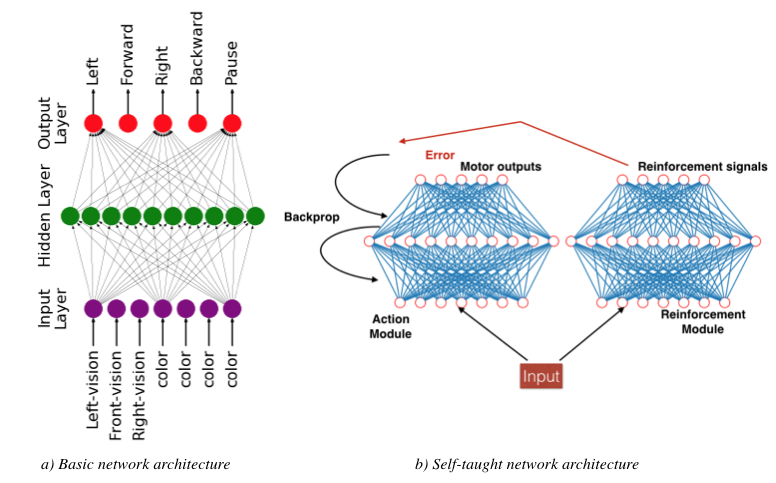}
\caption{Neural network controller for each situated agent.
\textbf{a)} Basic network without self-supervised learning ability. Each neural network includes 3 layers with 7 input nodes in input layer, 10 nodes in hidden layer, and 5 nodes in output layers. The first layer takes as input what an agent senses from the environment in its visual range (described above). The output layer produces 5 values in which the max value is chosen as a motor-guidance. \textbf{b)} Self-supervised neural architecture. The difference between the output of the reinforcement module and the action module is used to update the weights in action modules through Backpropagation. Through this self-supervised learning process, the action module approximates its output activation towards the output of the reinforcement module. The learning rate is 0.01.}
\label{fig:neural-controller}
\end{center}
\end{figure}

To allow for self-supervision, the neural controller for each agent now has two modules, namely \textbf{Action Module} and \textbf{Reinforcement Module}. The action module is the same network as previously shown in Figure \ref{fig:neural-controller}a). This module takes as sensory inputs and produce motor outputs. The reinforcement module has the same set of inputs as the action module, but has separate hidden and output neurons\footnote{For simplicity, I use the same topology for both modules}. The goal of reinforcement network is to provide \textit{reinforcement signals} to guide the behaviour of each agent through a self-supervised learning process. This architecture allows for an \textbf{intrinsic} learning process within the agent itself, no need for external supervision provided by the experimenter or reward like in DRL. Moreover, unlike \textit{actor-critic DRL} the reinforcement module here does not provide reward signal and memory to update the action network, and it is not updated by changing the role of two networks \cite{Mnih2015}. We shall see it is evolution that creates the self-supervision ability and the reinforcement signals, to make the evolved agent learns better. The description of the self-supervised neural network is visualised in Figure \ref{fig:neural-controller}b). Simulations are descried below.

\subsection{Simulation 1: Evolution alone (EVO)}
I use an evolutionary algorithm to evolve the population of weight matrices, one of the most common approaches to neuroevolution \cite{Stanley2019}.

The multi-agent system goes through a total of 100 generations, with 5000 time steps per generation. At each time step, an agent does the following activities: Perceiving MiniWorld through its sensors, computing its motor outputs by its neural network, moving in the environment which then updates its new heading and location. In evolution alone simulation, the agent cannot perform any kind of learning during its lifetime. After that, selection chooses individuals based on the number of food eaten in the foraging task used as the fitness value. The higher the number of food eaten, the higher the fitness value. For crossover, two individuals are selected as parents, namely $p_1$ and $p_2$, in order to produce one offspring called $child$. Crossover is implemented as the more successful parent has a higher the chance of passing its weights\footnote{Suppose \textbf{ratio} = $f_1$/($f_1+f_2$) (where $f_1$, $f_2$ = the fitness of $p_1$ and $p_2$, respectively). For each weight $w$ in the weight matrix of the child, we role a random number $rand$. If $rand < ratio$,  w is copied from the same corresponding weight in $p1$; otherwise from $p2$}.

Once a child has been created, that child will be mutated based on a predefined \textit{mutation rate} 0.05\footnote{If mutation occurs for each weight in the child, that weight is added by a random number from the range [-0.05, 0.05] -- a \textit{slight mutation}}. After that, the newly born individual is placed in a new population. This process is repeated until the whole new 20 agents are created.

\subsection{Simulation 2: Evolving Self-supervised agents}
In this simulation, I allow lifetime learning, in addition to the evolutionary algorithm, to update the weights of neural network controllers. I evolve a population of \textbf{Self-supervised} agents -- agents that can supervise, or teach themselves. During the lifetime of an agent, the reinforcement modules produce outputs in order to guide the weight-updating process of the action module. Only the weights of action modules can be changed by learning, the weights of reinforcement module are genetically specified in the same evolutionary process as specified above in Evolution alone.

%We can interpret this scenario as an agent has an ability to produce reinforcement signals to guide itself. It is evolution that produces these reinforcement signals, or the desire to \textit{external stimuli} (the sensory inputs in this case), for every agent. In other words, it is evolution that provides the self-taught ability for each agent. This is how evolution influences learning. And more than this, it is learning during the lifetime that changes the fitness of each agent, hence the fitness landscape which then affects the evolutionary process. This is the interaction between learning and evolution which is being investigated in this paper. Due to scope of this paper, we focus more on the influence of learning on the performance of the evolving multi-agent system in MiniWorld. 

I use the same parameter setting for evolution as in EVO simulation above. At each time step, an agent does the following activities: Receiving sensory inputs, computing its output, updating its new heading and location, and updating the weights in action module by \textbf{self-teaching}. After 5000 steps per generation, the multi-agent system undergoes selection and reproduction processes as in Evolution alone\footnote{Remember that we are fitting self-learning and evolution in a Darwinian framework, not Lamarckian. This means what will be learned during the lifetime of an agent (the weights in action module) is not directly passed down onto the offspring}.

\subsection{Simulation 3: Self-supervised agents Alone}
I conduct another simulation in which all agents are self-taught agents -- having self-taught networks that can teach themselves during lifetime. What differs from simulation 2 is that at the beginning of every generation, all weights are randomly initialised, rather than updated by an evolutionary algorithm like in simulation 1. The learning agents here are initialised as \textit{blank-slates}, or \textit{tabula rasa}, having no predisposition or some sort of \textit{priori knowledge} about the world before learning. We are curious whether evolution brings any benefit to learning in MiniWorld.%In other words, we would like to see if there is a synergy between evolution and learning, not just how learning can affect evolution.

\section{Results, Analysis, and Discussion}
We compare the best and the average energy (fitness) absorbed by the multi-agent system in MiniWorld. All results are averaged over 30 independent runs. 

Before going into the result, through the visualisation of world maps and the description of what the agent has equipped, we can arrange world map A-B-C-D in the increasing order of difficulty to forage. In map A, all agents are initially born with a more tendency toward the food. The scenario is reversed in map B when the agent is born with a more tendency toward the poison. Map C and map D may present an even more difficulty for our game since the agent is born with no tendency to experience anything in the world\footnote{This is an important point because as described above, our agents are provided little information about the world, and they are \textbf{autonomous} and \textbf{situated} -- what they sense can affect their movement in the world, which then affect their future behaviour, so on and so on. Without good movement they may not able to sense more relevant information which then helps produce more relevant movement in the world.}. In fact, to be effective the agent has to learn two tasks: to sense the world correctly to find the substance, and from that develop the ability to discriminate between food and poison, which will be used to guide its movements.

\subsection{Self-supervised Learning Facilitates Evolution}

First we would like to see how self-supervised learning facilitates the evolutionary search by comparing EVO+Self-supervised (EVO+SS) and EVO alone. A similar trend can be observed in nearly all experimental settings is that there is a decrease in performance from A to B to C to D. This is understandable because of the increasing difficulty to forage in these maps as just analysed above.

\begin{figure}[t]
\begin{center}
\includegraphics[width=5in]{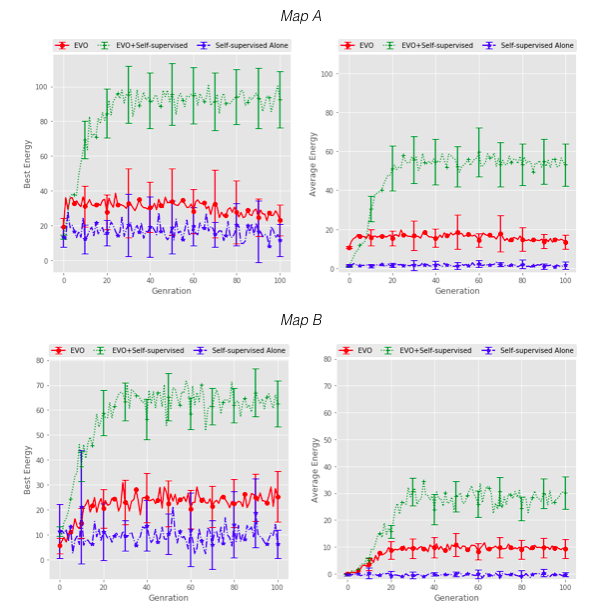}
\caption{Energy comparison. Top=Map A, Bottom = Map B.}
\label{fig:energy1}
\end{center}
\end{figure}

How each type of experimental setup performs compared to each other? It can be seen that EVO+SS outperforms EVO alone in all maps with respect to both the best and the average energy. Specifically, in the easiest case (map A) we see that the best agent in EVO+SS, on average, is around 60 energy higher the best agent in EVO alone. Overall, the whole EVO+SS system absorbs 800 (40 on average) energy higher than the EVO alone. In map B, a harder case, we that the difference between these two systems (EVO and EVO+SS) is reduced, yet EVO+SS still has a clearly better performance. %Overall, the whole EVO+SS system absorbs 400 (20 on average) energy higher than the EVO alone. 

\begin{figure}[t]
\begin{center}
\includegraphics[width=5in]{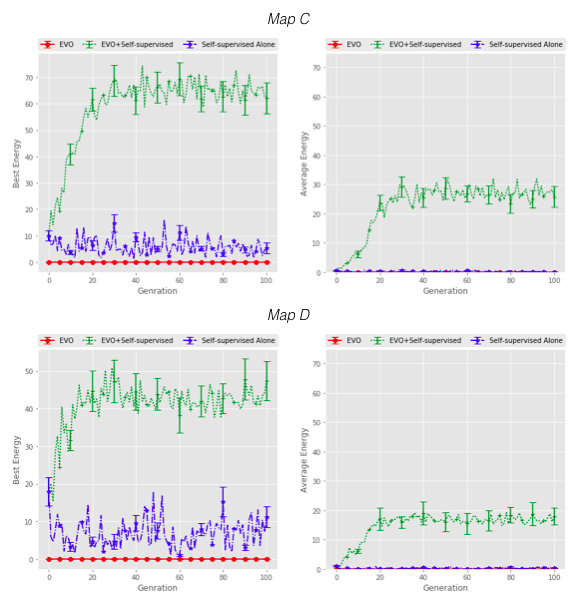}
\caption{Energy comparison. Top=Map C, Bottom = Map D.}
\label{fig:energy2}
\end{center}
\end{figure}

Interestingly, in map C and D the EVO alone system cannot eat at all. This is predictable. Simply speaking, an agent that cannot learn can only use its \textit{innate} ability, hardwired in its brain, to search the environment. However, as analysed above, the agent in EVO alone is born without any tendency to sense relevant information (about the substance) in the environment, and also has no ability to change its motor program hardwired in its brain. This is why they cannot eat anything at all since their actions are repeatedly making no sense in MiniWorld. Conversely, the system of self-supervised agents still can absorb energy in maps C and D. 

\begin{figure}[t]
\begin{center}
\includegraphics[width=5in]{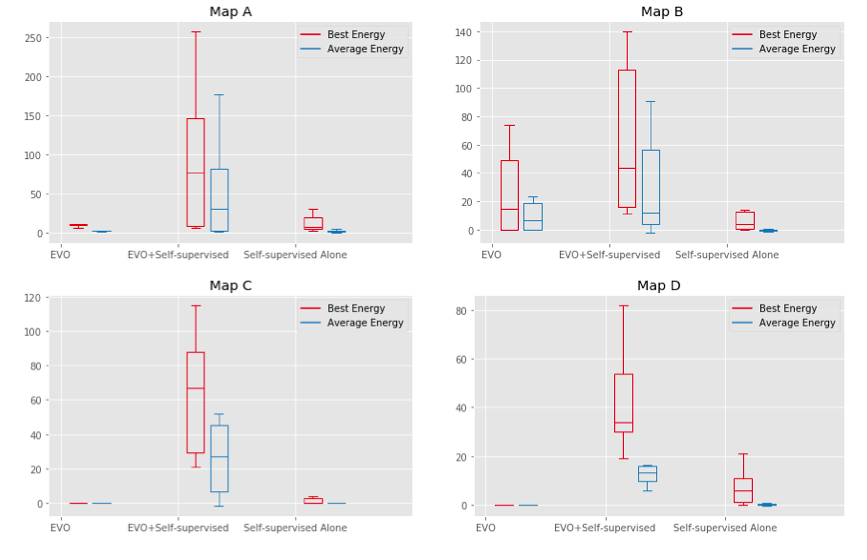}
\caption{Boxplot comparison in terms of energy between systems.}
\label{fig:boxplot-energy}
\end{center}
\end{figure}

Moreover, the boxplots in Figure \ref{fig:boxplot-energy} again show that EVO+SS outperforms EVO alone in all setups, and that this difference is statistically significant. 

Not just the performance, one curious question arising here is why self-supervised agents can behave that way? One plausible explanation for this is the effect of self-supervised learning on evolution: By doing self-supervised learning, the EVO+SS system can facilitate the evolutionary search. This dominance helps show that the agent in EVO+SS may have successfully developed the ability to sense the world correctly and the ability to discriminate between food and poison, so that they can eat more food and less poison.

\subsection{Evolution Facilitates Self-supervised Learning}

We conduct the third simulation in which the neural networks controlling self-supervised agents are all randomly initialised at the beginning of each generation, without the participation of evolution. It can be observed in Figure \ref{fig:energy1} and Figure \ref{fig:energy2} that in all maps, the Self-supervised alone (SS) system has a clearly lower performance than EVO+SS, especially when it comes to the harder cases (map C and map D). The difference is statistically significant as shown in Figure \ref{fig:boxplot-energy}. It is also interesting that in our simulations, the best \textbf{blank-slate} SS agent cannot outperform the best in EVO alone in easier cases (map A and B), but has some small advantage in harder cases (map C and D) when the agent in EVO alone cannot search for any food at all.

Another thing to be seen here is the average energy of the SS system is relatively close to 0 while the best agent still can absorb a positive amount of energy in all maps. This behaviour is interesting because it shows that through self-supervised learning some agent can develop a relatively effective behaviour to both explore the world ((even if it is born with tendency to reach nothing in map C and D), and behave as if it can eat more food than poison. This also indicates that the whole SS system without evolution cannot successfully develop the ability to discriminate between food and poison, hence the average energy is zero. 

This can be explained as with the ability to teach oneself by leveraging the difference between the action and the reinforcement modules, the weights of the action module of some agent may have been changed to produce better movements. The agent that can reach the food source has a higher chance of being selected to leave offspring. Thus, its \textit{good} genetic information, consisting of the \textit{initial} weights of both the action and the reinforcement modules, is more likely to proliferate. When an agent is selected for reproduction, its self-supervising, or self-teaching, ability is likely to be also promoted in later generations. It is this ability that has made future evolved self-supervised agents better at teaching themselves in order to develop more effective movement in MiniWorld. This process repeats as what has given advantage during lifetime of the self-taught agent is preserved and promoted by the evolutionary process.

%The dominance of EVO+SS over SS alone shows how evolutionary search has facilitated the self-supervised learning process, making better self-supervised neural networks over time to help agents experience the world, and from that develop more correct behaviour when faced with food and poison substances. Not just the performance, one curious question arising here is why EVO+SS can evolve more adaptive behaviour in our world, especially in map C and D? We are going to briefly explain this question.

It is plausible conclude that some sort-of intelligent foraging behaviour has emerged in our autonomous multi-agent system, through the interaction between evolution and self-learning, and this is what I have observed in my simulation\footnote{The video demonstration has been made anonymously in this link https://youtu.be/aUlLJfisAHA.}. Through self-taught learning, some initial advantage behaviour would have been emerged, and that behaviour changed the evolutionary pathway, which further that adaptive behaviour in the future. This is, through the evolutionary process, some priori-knowledge about the environment can be encoded in the neural network controller. This makes agents having predisposition to learn adaptive behaviours learn more effectively than blank-slate agents. This is the interplay between evolution and self-supervised learning.

%It is plausible here to conclude that self-supervising can provide more adaptive advantage than the slower evolutionary process in MiniWorld. However, it is evolution that provides a good base for self-taught agents to learn better adaptive behaviours in future generations rather than learning as \textit{blank-slates} in Self-supervised alone population. This can also be explained by the understanding of the synergy between evolution and learning. Through the evolutionary process, some priori-knowledge about the environment can be encoded in the neural networks controlling agents. Agents having priori-knowledge, or predisposition to learn adaptive behaviours in our scenario, can learn faster and learn more adaptively than blank-slate agents. This is the Baldwin-like Effect -- the interplay between learning and evolution.

\section{Conclusion and Further Discussion}
I have presented and analysed a technique called Evolving Self-supervised Neural Networks. Behind this technique is the interplay between evolution and learning in the form of self-supervision in order to produce ever more adaptive and intelligent behaviour. This proposed network has been used to control a situated autonomous multi-agent system in which the agent requires very little domain and human-engineered knowledge. Experimental results in increasingly difficult maps have shown that the Evolved Self-supervised system outperforms both the EVO and the Self-supervised learning system alone. The interplay between self-learning and evolution has also been demonstrated in the sense that evolution facilitates future self-supervised learning, better than learning as \textit{blank-slates}. Through this interaction, an there is an emergence of intelligent behaviour in our system. 

Here I stress more onto the philosophical idea that self-learning and evolution can interact and benefit each other in evolving intelligence. This differs from traditional point of view in EAs, and also the blank-slate point of view in that here we have evolved a learner better at learning to solve a problem with more innate information gradually encoded by evolution, rather than having evolved a solution to that problem. This fits with the current understanding of human cognition and intelligent behaviour in cognitive science and evolutionary epistemology. It seems to open another way to reach human-like intelligence in the computer: \textit{Not evolve to solve, not learn as blank-slate, but evolve a learner to solve}.

The computational method is simple enough to illustrate the idea, but still has some indications. The idea of self-taught neural networks can be powerful when there is no external supervision (or \textit{label} provided from external data). This opens a way to create intelligent autonomous multi-agent learning, in hopes of producing autonomous intelligence \cite{Stone:2007}, \cite{OpenAI-Multiagent}. The algorithm and technique used in this paper can also be a potential technique to solve unsupervised learning, or learning with limited label data (weak supervision) \cite{Zhou2017}, especially in reinforcement learning and games \cite{sutton2018reinforcement}. We are curious whether evolution can provide a better base to learn than learning as blank-slates like what was claimed by DeepMind in games \cite{Mnih2015}. Indeed, the shallow network used in this paper does not restrict the application of the core philosophical idea into deep neural networks, as long as we can combine evolutionary search and the idea of self-taught neural architecture by employing variants of gradient-based learning.

There is some limitation that should not be neglected, including the use of a fixed neural architecture. This limitation has also been raised in a few in current deep learning research \cite{Schmidhuber14}, \cite{Stanley2019}. One plausible solution could be evolving both the weights and the topology of a neural network \cite{Stanley:2002:NEAT}, \cite{UberAI-Stanley-DeepNE}. This is an interesting pathway for future work if we can evolve variable self-supervised neural architecture which can be an intrinsically general neural learner.

%\bibliographystyle{unsrt}  
%\bibliography{arxiv.bib}


\end{document}